# Title

Behavioral Safety Assessment towards Large-scale Deployment of Autonomous Vehicles

# Authors

Henry X. Liu[1,2,#,*], Xintao Yan[1,#], Haowei Sun[1,#], Tinghan Wang[1,#], Zhijie Qiao[1], Haojie Zhu[1], Shengyin Shen[2], Shuo Feng[2], Greg Stevens[2], Greg McGuire[2]

# Affiliations

[1]Department of Civil and Environmental Engineering, University of Michigan, Ann Arbor, MI, USA
[2]University of Michigan Transportation Research Institute, University of Michigan, Ann Arbor, MI, USA
[#]Equal Contribution
*Corresponding Author, henryliu@umich.edu

# Abstract

Autonomous vehicles (AVs) have significantly advanced in real-world deployment in recent years, yet safety continues to be a critical barrier to widespread adoption. Traditional functional safety approaches, which primarily verify the reliability, robustness, and adequacy of AV hardware and software systems from a vehicle-centric perspective, do not sufficiently address the AV's broader interactions and behavioral impact on the surrounding traffic environment. To overcome this limitation, we propose a paradigm shift toward behavioral safety—a comprehensive approach focused on evaluating AV responses and interactions within the traffic environment. To systematically assess behavioral safety, we introduce a third-party AV safety assessment framework comprising two complementary evaluation components: the Driver Licensing Test and the Driving Intelligence Test. The Driver Licensing Test evaluates the AV's reactive behaviors under controlled scenarios, ensuring basic behavioral competency. In contrast, the Driving Intelligence Test assesses the AV's interactive behaviors within naturalistic traffic conditions, quantifying the frequency of safety-critical events to deliver statistically meaningful safety metrics before large-scale deployment. We validated our proposed framework using Autoware.Universe, an open-source Level 4 Automated Driving System (ADS), tested both in simulated environments and on the physical test track at the University of Michigan's Mcity Testing Facility. The results indicate that Autoware.Universe passed 6 out of 14 scenarios and exhibited a crash rate of $3.01 \times 10^{-3}$ crashes per mile—approximately 1,000 times higher than the average human driver crash rate. During the tests, we also uncovered a number of unknown unsafe scenarios for Autoware.Universe. These findings underscore the necessity of behavioral safety evaluations for improving AV safety performance prior to widespread public deployment.

# Introduction

Autonomous vehicle (AV) technology is transforming modern transportation, promising increased safety, efficiency, and accessibility. In recent years, AV deployments have expanded globally. Waymo, for example, offers 24/7 paid services in Phoenix, San Francisco, and Los Angeles[1], and is planning to expand internationally to cities such as Tokyo. Similarly, Baidu's Apollo Go has launched operations in Wuhan and other Chinese cities[2]. However, these deployments have sparked widespread debate, particularly regarding AV safety[3]. High-profile incidents, such as when a Cruise robotaxi in 2023 hit and dragged a pedestrian about 20 feet in San Francisco[4], have intensified public scrutiny and eroded trust in AV technology. A recent survey shows that only 13% of U.S. drivers trust riding in AVs[5], highlighting the public's skepticism and the pressing need to build confidence. Public acceptance and consumer confidence depend heavily on perceived safety[6], making unbiased, rigorous, transparent, and publicly accessible safety evaluation programs essential. Furthermore, systematic AV safety assessment frameworks are crucial for regulators to establish effective safety standards and facilitate the seamless integration of AVs into existing transportation systems.

Despite the critical need, a systematic AV testing program—particularly through third-party evaluations—remains absent. While well-established testing programs exist for conventional human-driven vehicles[7-10], they are inadequate for Level 4-5 AVs. For example, the New Car Assessment Program (NCAP) is a safety rating program that primarily assesses vehicle crashworthiness and crash avoidance technologies such as Autonomous Emergency Braking (AEB). For high-level AVs, notable standards include ISO 26262[11] for Functional Safety (FuSa), ISO 21448[12] for Safety of the Intended Functionality (SOTIF), and ISO/PAS 8800[13] for Road Vehicles Safety and Artificial Intelligence, all of which address risks arising from hardware and software malfunctions or insufficiencies. In addition, ISO 34502[14] introduces a scenario-based testing framework for SOTIF, while UL 4600[15] provides methods for assessing AV's overall safety case. However, these testing standards adopt a *vehicle-centric perspective*, assessing AV safety based on vehicle system function and capability, such as sensor failures (hardware) and perception errors (software). While essential, a vehicle-centric approach is insufficient for evaluating AV safety in large-scale real-world deployments for two key reasons. First, large-scale deployment requires evaluating AVs' safety impact on the surrounding traffic environment, including not only their ability to mitigate or avoid hazards but also their capacity to avoid creating new safety hazards. This necessitates an *environment-based* approach that evaluates how AVs react to and interact with dynamic agents in stochastic, diverse, and complex traffic situations, an aspect FuSa and SOTIF do not address. Second, AV testing must shift from *deterministic assessment* to *probabilistic evaluation*, as modern AVs rely heavily on deep learning models that exhibit inherently non-deterministic behavior. In other words, safety evaluation based solely on predetermined scenarios is insufficient; unbiased estimation of the overall crash rate is necessary to justify the net societal benefits of AVs and to prevent damage to consumer trust from singular unreasonable risks. It also helps answer the fundamental question, "How safe is safe enough?"

The above discussions underscore the necessity of shifting the AV evaluation paradigm from *functional safety* to **behavioral safety**. Behavioral safety, as defined herein, refers to the *absence of unreasonable risk due to hazards resulting from an AV's unsafe behavioral responses to and interaction with the traffic environment.* Specifically, it encompasses two key aspects of unsafe AV decision-making: (1) reactive behavior, which involves failures in avoiding or mitigating hazards when presented with safety-critical situations, and (2) interactive behavior, which concerns both an AV's potential to unintentionally expose itself to safety-critical situations and its inability to minimize the likelihood of secondary incidents. To illustrate behavioral safety, consider a scenario where a human-driven background vehicle (BV) aggressively cuts into the AV's lane. Even if the AV system is fully functional and detects the maneuver accurately, behavioral safety evaluation assesses the AV's capability to make appropriate decisions—such as timely braking or performing an evasive lane change—without causing another safety-critical situation. Behavioral safety also includes proactive risk mitigation measures, such as avoiding prolonged driving in another vehicle's blind spot to reduce potential cut-in incidents. Furthermore, behavioral safety explicitly evaluates the repercussions of an AV's interactive behavioral decisions in preventing secondary incidents. For instance, when an AV executes a hard braking maneuver, behavioral safety assessment must consider whether this decision could inadvertently cause a rear-end collision with a following BV. Ultimately, behavioral safety evaluates whether an AV can intelligently *react to* and *interact with* both static road infrastructure and dynamic road users, when the *system is functionally sufficient and free of errors*. Consequently, behavioral safety is fundamentally distinct from crashworthiness and functional safety, as shown in Fig. 1a. The CertiCAV Assurance Paper[16] by Connected Places Catapult of the United Kingdom also highlights the importance of behavioral safety, reinforcing our argument that functional safety alone is insufficient for AV evaluation. However, it falls short of providing concrete definitions and practical methodologies for implementation, leaving a critical gap in the current AV testing approaches. In recent years, while some AV companies have released information about their internal testing procedures[17-19] related to behavioral safety, these disclosures remain limited, and the specifics of their testing protocols are often inaccessible. Therefore, developing a systematic, transparent, and robust third-party testing framework for behavioral safety of AVs is an urgent priority.

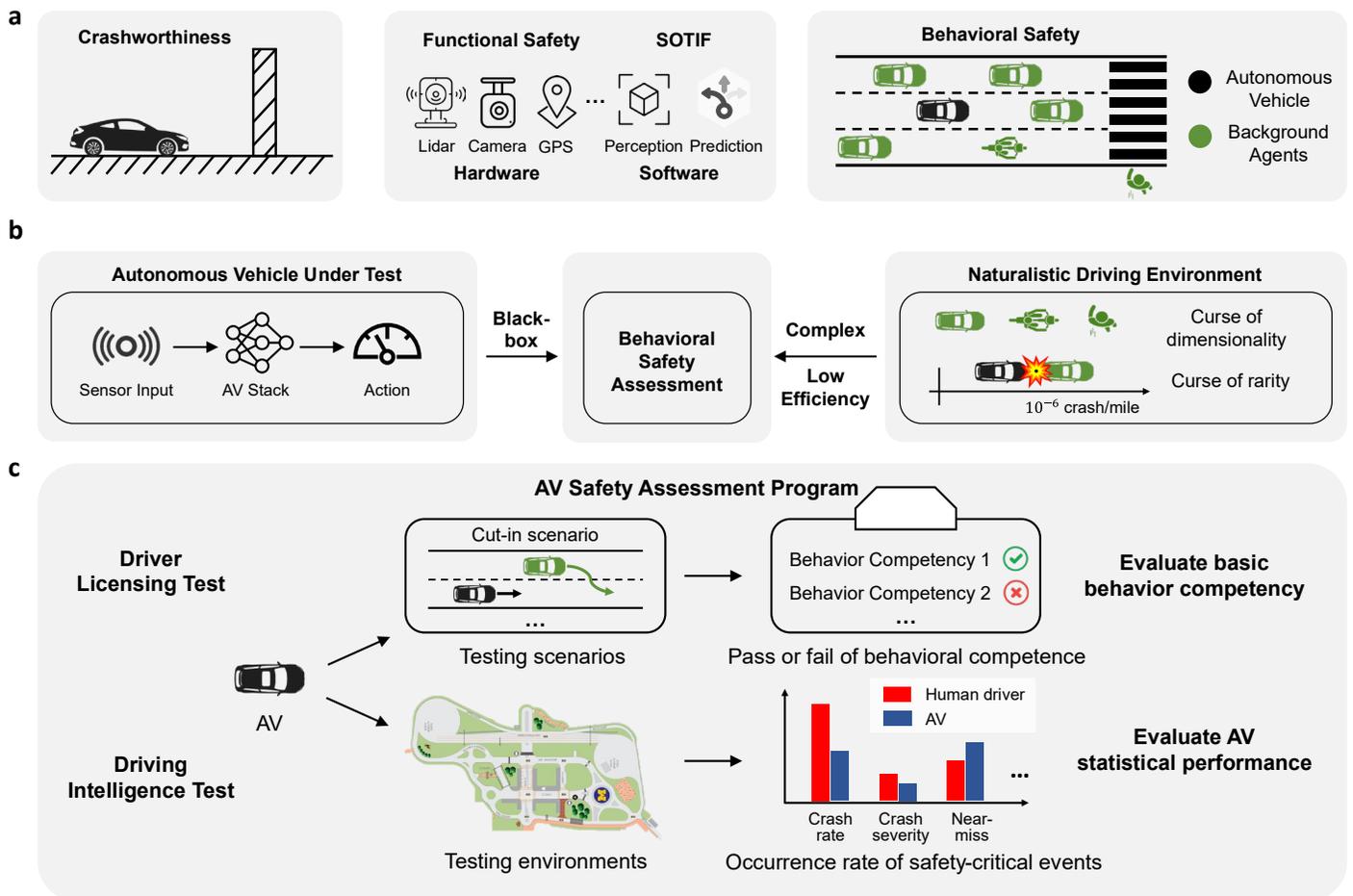

**Figure 1. Demonstration of behavioral safety assessment for AV. a,** Illustration of the differences between behavioral safety and existing testing methods. **b,** Major challenges for AV behavioral safety evaluation. **c,** Illustration of the proposed framework.

Conducting third-party evaluations of AV behavioral safety presents significant challenges, as illustrated in Fig. 1b. First, modern AVs—particularly end-to-end models—lack interpretability and operate as black-box systems, making failures unpredictable. This means failures may occur not only in rare edge cases but also during routine driving. Such unpredictability erodes public trust, especially when crashes happen under normal conditions. A recent study shows that over 90% of AV crashes occur in situations with no unusual road conditions or dangerous pre-accident behaviors[20], contradicting the common belief that AV testing should focus primarily on rare edge cases. This highlights the need for a holistic evaluation of basic behavioral competency, rather than just extreme-case testing. Second, AV testing is inherently difficult due to the complexity of real-world traffic and the rarity of safety-critical events. The naturalistic driving environment (NDE) involves intricate spatial-temporal interactions among diverse road users, across varying road geometries, and under dynamic traffic conditions[21]. Accurately replicating these factors in simulations requires high-fidelity modeling of both physical environments and human behaviors, making AV assessment highly complex. Compounding this challenge is the curse of rarity[22]—safety-critical events occur infrequently, requiring billions of miles to collect sufficient data for meaningful evaluation[23]. This renders brute-force mileage-based testing impractical, even in simulations, where the key challenge is generating safety-critical events both realistically and efficiently.

To address these challenges, we propose a dedicated third-party testing initiative for systematically evaluating AV behavioral safety. The framework consists of two key components: the **Driver Licensing Test (DLT)** and the **Driving Intelligence Test (DIT)**, as illustrated in Fig. 1c. The DLT focuses on evaluating AV performance in *"known"* scenarios, assessing basic behavioral competency based on human prior knowledge. Inspired by human driver's license tests, it examines fundamental driving skills across both normal and safety-critical situations, such as car-following and cut-ins. Each scenario includes multiple risk levels, with adversarial background agents introduced in high-risk cases to provoke challenging situations, enabling assessment of the AV's *reactive behavior* in avoiding and mitigating hazards. This approach enables a broad evaluation beyond rare edge cases, providing a baseline assessment of AV behavioral safety in situations where human drivers are expected to perform reliably. It conceptually aligns with existing scenario-based testing practices commonly adopted by testing organizations[24] and AV companies[25].

In contrast, the DIT uncovers *"unknown unsafe"* events through statistical evaluation, providing a comprehensive assessment of AV behavioral safety within its Operational Design Domain (ODD) by considering both *reactive and interactive behaviors*. Instead of relying on predefined scenarios, the DIT constructs a high-fidelity NDE to create a realistic background traffic environment for AV assessment. To overcome the curse of rarity, we introduce the Naturalistic and Adversarial Driving Environment (NADE), which builds upon NDE to intelligently increase the exposure frequency of safety-critical events using importance sampling theory. This enables the DIT to measure the occurrence rate of safety-critical events (e.g., crashes) involving both AVs and BVs, accelerating testing without introducing bias, thus significantly improving efficiency while maintaining statistical rigor. Beyond statistical performance measurement, a key advantage of the DIT is its ability to reveal previously unknown failure modes, which is particularly crucial before large-scale AV deployments. Together, DLT and DIT form a complementary framework, addressing both known and unknown sources of AV unpredictability. The DLT ensures AVs possess essential behavioral competencies expected of human drivers, while the DIT identifies long-tail risks that may only emerge at scale. A key strength of our framework is its testing-oriented design, which treats the AV as a black box, ensuring fairness and objectivity by not requiring prior knowledge of its internal mechanisms.

To demonstrate the effectiveness of our approach, we applied it to evaluate a Level 4 AV system running open-source automated driving software, Autoware.Universe[39], in a simulated testing environment at the University of Michigan's Mcity Test Facility. We assessed the AV's performance within an urban driving ODD, including intersections, roundabouts, and urban arterials, under normal weather conditions. High-fidelity testing scenarios and environments were constructed using large-scale real-world datasets that model human driving behavior. For the DLT, we selected 14 representative scenarios—comprising both single-agent and multi-agent scenarios—that represent 71.5% of all crashes within the ODD[26] to evaluate the behavioral competency. The DIT was developed based on state-of-the-art intelligent testing methodologies[27] to obtain statistical performance of the AV accurately and efficiently. Simulation results reveal that Autoware.Universe, the latest version of Autoware, passes eight out of 14 scenarios in the DLT. The DIT estimated a crash rate of $3.01 \times 10^{-3}$ crashes per mile, about 1,000 times higher than that of an average human driver[38] ($3.00 \times 10^{-6}$ crash per mile). These testing results successfully identified unknown unsafe events, exposing potential weaknesses and bugs in the AV's software stack, such as its failure to navigate certain roundabouts, leading to unexpected stops. This issue was reported to the developers, who were able to address and resolve it based on our feedback. These findings highlight the framework's effectiveness in detecting and diagnosing system vulnerabilities, particularly those unknown unsafe behavior which is essential for refining AV systems and supporting a closed-loop development process for large-scale deployment. To further validate the applicability of our framework in real-world settings, we conducted physical tests using Mcity's closed test tracks with a Tesla Model 3 (equipped with Full Self-Driving Version 12.5.5) and a prototyping vehicle equipped with Autoware.Universe on a Lincoln MKZ platform. Due to the resource constraints, the goal was not comprehensive testing but to demonstrate feasibility. For the DLT, we tested two scenarios with the Tesla—one involving a vehicle and the other involving a vulnerable road user (VRU)—both of which were successfully passed across multiple risk levels. For the DIT, testing with the Autoware-equipped Lincoln MKZ revealed that the system could benefit from further improvements to proactively reduce risk and limit exposure to safety-critical situations. These results highlight the framework's versatility and effectiveness in evaluating AV systems across both simulated and physical environments, reinforcing its potential as a systematic, standardized, and third-party safety validation tool for AV deployment.

## Results

### Behavioral safety assessment for autonomous vehicle

The design process for both the DLT and DIT follows a unified framework, as illustrated in Figure 2. The first step involves creating a NDE based on Naturalistic Driving Data (NDD) and the specific ODD. The NDE forms the foundation for behavioral assessment, simulating real-world traffic conditions. However, directly generating and sampling testing environments from the NDE can be inefficient[23,29], as safety-critical events are rare in real-world traffic[22], rendering many environments irrelevant for safety assessments. To address this, we introduce the NADE, which builds upon the NDE to create and sample testing scenarios and environments more effectively. The key idea of the NADE is to intensify the occurring frequencies of safety-critical events while maintaining comprehensive coverage for the DLT and ensuring unbiased results in the DIT. It is important to note that the difference between a "scenario" and an "environment" largely pertains to scale and complexity: environments encompass broader spatiotemporal coverage, with a larger number of background road users. Various scenarios, featuring different types of interactions, can emerge spontaneously and dynamically within the environment. In this sense, a scenario can be viewed as a simplified version of an environment, which is why we refer to the term "environment" in the NDE and NADE throughout this paper.

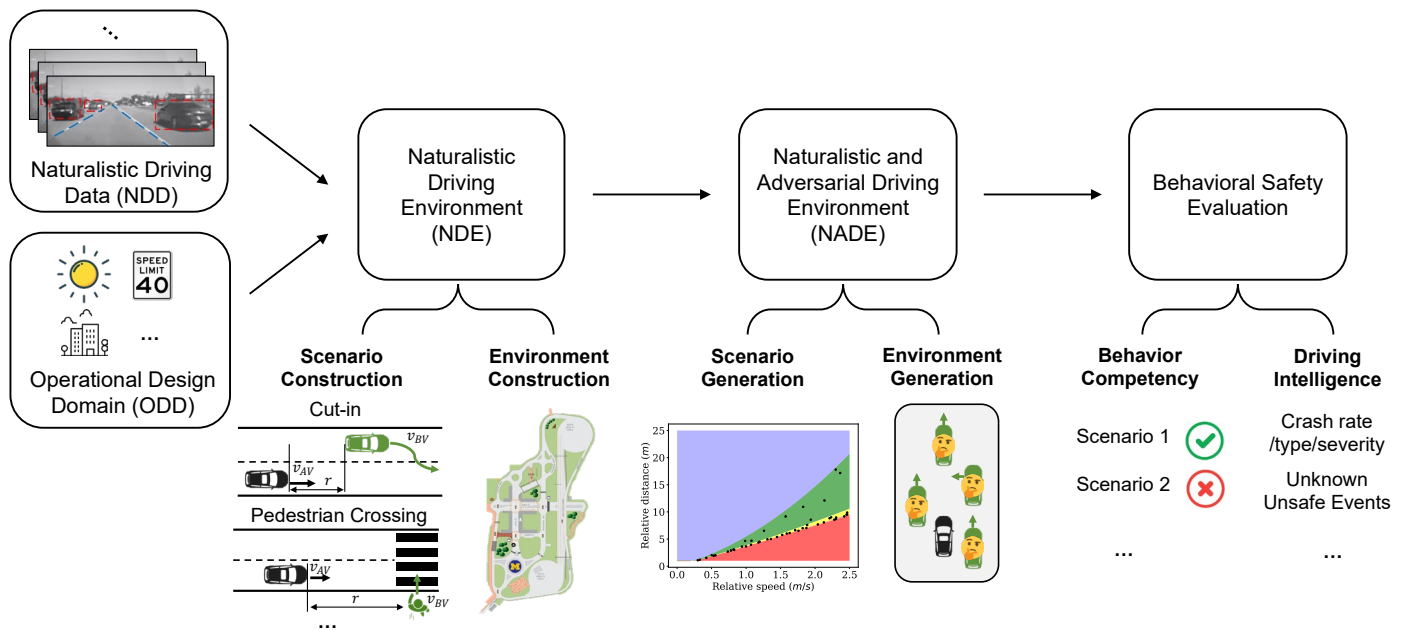

**Figure 2. Overall pipeline of developing the *Driver Licensing Test* and the *Driving Intelligence Test*.**

### Development process of the Driver Licensing Test

This section details the development process of the DLT. The initial step involves abstracting basic behavioral competencies and building a testing scenario library grounded in real-world dynamic driving tasks. Several existing studies, standards, and best practices contribute to the creation of scenario libraries[30-34]. Based on this

Page 5 of 25

foundation, we compiled our testing scenario set, considering the typical driving scenarios within the urban ODD. Specifically, we leveraged the scenario library from Ref[34] to guide our selection and curated 14 representative scenarios, as shown in Fig.3a. In these scenarios, the black vehicle represents the AV under test, the white vehicle represents the BV, and the human figure represents the VRU. The selected scenarios encompass both single-agent situations, such as the traffic signal scenario, as well as multi-agent scenarios featuring various road user types. Notably, these scenarios encompass 71.5% of crash types observed in real-world urban driving environments[26], offering reasonable coverage. While the selected 14 scenarios may not fully represent the entirety of the urban ODD, they were chosen to demonstrate our proposed methodology within the constraints of resources. The methodology is designed to be flexible, allowing for the addition of more scenarios as needed.

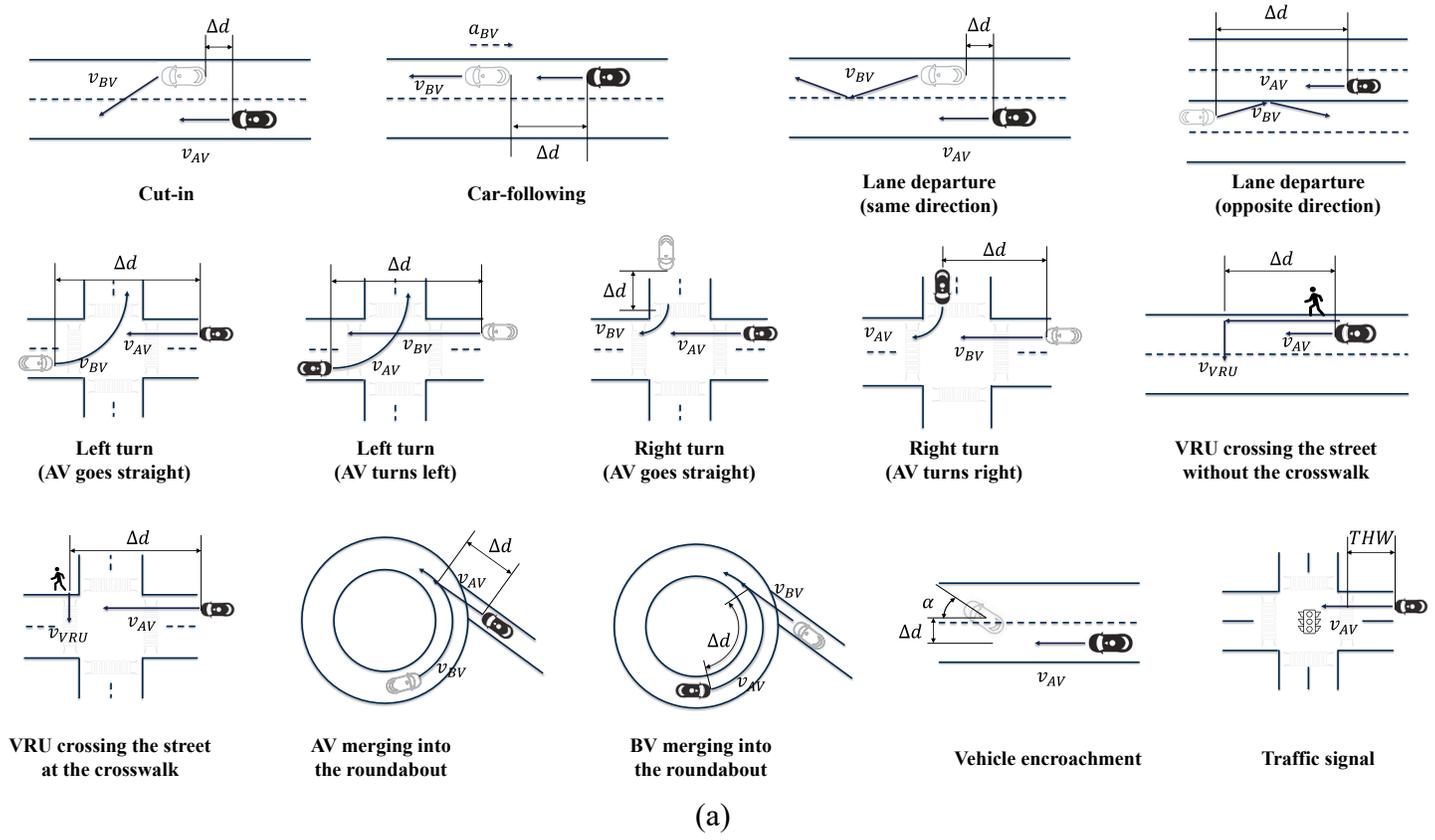

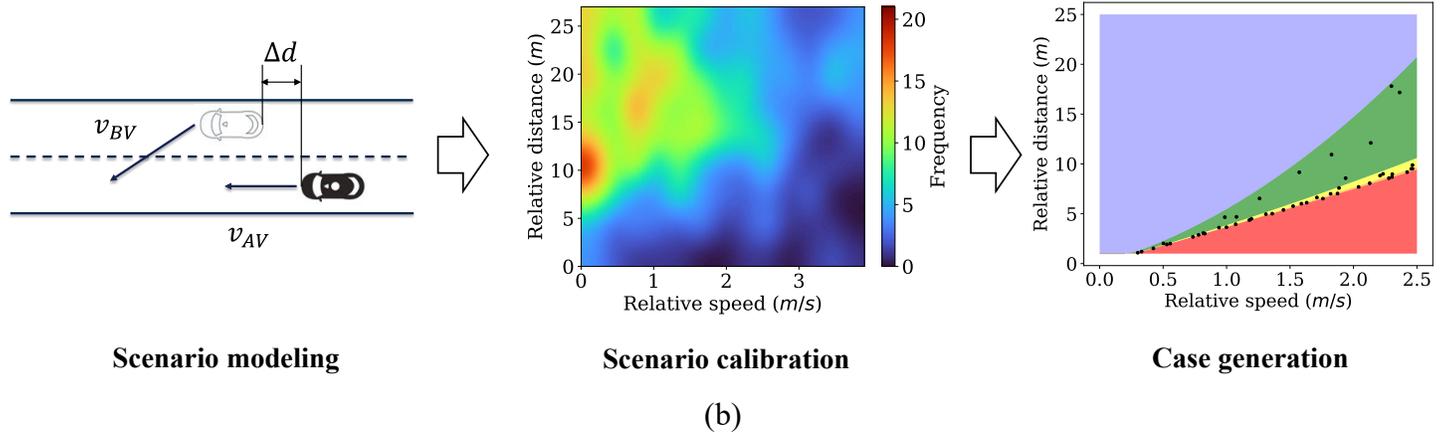

**Figure 3. Demonstration of the development process of the Driver Licensing Test. a,** Demonstration of the scenario set used in the Driver Licensing Test of this study. The black vehicle represents the AV under test, the white vehicle represents the background vehicle (BV), and the human figure represents the vulnerable road user (VRU). **b,** Demonstration of the cut-in scenario development pipeline.

Each scenario follows a three-step development process: scenario modeling (parameterization in state space), scenario calibration (constructing the scenario's NDE distribution), and case generation (building the NADE distribution for test case sampling). As shown in Fig.3b, we demonstrate the three-step process using a cut-in scenario. Since the AV is treated as a black box, the tester cannot control its speed during testing. Therefore, for a cut-in scenario, the dynamic variables are the relative speed and relative longitudinal distance between the AV and the BV, and the BV will execute lane changes at a constant speed, following a sine-function trajectory. As a result, the cut-in scenario is modeled as a two-dimensional problem, where each test case is uniquely defined by the initial relative speed and longitudinal distance between the AV and the BV.

In the scenario calibration step, we aim to derive real-world distributions of scenario parameters. Using the Argoverse 2 Motion Forecasting Dataset[35], we extracted 11701 cut-in segments and developed the NDE distribution for this scenario, as shown in Fig.3b (middle). Rather than randomly sampling from the naturalistic distribution, we balance coverage and difficulty in the case generation process to account for different risk levels. For each test case, we calculate the minimum deceleration required by the AV to avoid a collision. By referencing human drivers' deceleration distributions in cut-in events from the NDD, we categorize the scenario state space into five risk levels—trivial, low, mid, high, and infeasible—based on the required deceleration. These levels are color-coded in Fig.3b (right): blue (trivial), green (low), yellow (mid), orange (high), and red (infeasible), corresponding to different percentiles of human deceleration: above 80%, 80%-10%, 10%-1%, below 1%, and beyond observed human capabilities. For example, if the required deceleration falls between the 1st and 10th percentiles of the cut-in event data, the test case is categorized as mid-risk. We will generate test cases from low, mid, and high-risk levels by discretizing the state space into cells, with a higher sampling density in high-risk regions, and each cell serving as a sampled test case. Fig.3b (right) illustrates the sampling results, where each black dot represents a test case. The number of generated test cases for all scenarios is presented in the second column of Table 1, where $N$ represents the generated case number. More development and implementation details of different scenarios can be found in Supplementary Materials. Each test case is run 3 times to account for stochastic variability in the AV and testing systems. A test case is considered a failure if a collision occurs in any run, and a success if no collisions occur. By analyzing the results across different scenarios, we can assess the AV's basic behavioral competency. The AV is deemed to have failed the DLT if it encounters a failure in any of the testing scenarios. An illustration video of the DLT is provided in Supplementary Movie 1.

**Development process of the Driving Intelligence Test**

This section details the development process of DIT. Unlike the DLT, which models individual scenarios independently, the DIT begins by constructing a complete traffic environment through high-fidelity simulation of human driving behavior. This simulated NDE encompasses all relevant components of the AV's ODD, such as intersections and roundabouts, when evaluating performance in urban settings. The DIT enables the evaluation of an AV's capabilities over entire trips. To achieve this, we have developed *TeraSim*[36], an open-source, high-fidelity traffic simulation platform designed specifically for AV safety testing. TeraSim employs a generative paradigm to construct traffic environments using two core models. The NDE[21,37] reconstructs realistic traffic conditions with statistical fidelity, ensuring that generated environments follow real-world distributions, particularly for those safety-critical events. Building on NDE, the NADE[23,27] amplifies rare but critical events, systematically exposing AVs to safety-critical conditions to accelerate the evaluation of AV safety performance.

Our method enables high-fidelity simulation of the Mcity environment, which includes a 1,000-foot urban arterial, six signalized intersections (one with an unprotected left turn), three roundabouts, and various single and multi-lane roads, effectively replicating a realistic urban driving context. A key requirement for AV testing in the NDE is the accurate reproduction of safety-critical events based on real-world data[21,23]. To validate this, we compared the simulated crash rate in Mcity with ground-truth crash data from five years (2016–2021) in Michigan, US[38]. The ground-truth crash rate is $3.00 \times 10^{-6}$ crashes per mile, while our simulation yielded $2.93 \times 10^{-6}$ crashes per mile, confirming the fidelity of the NDE. Additionally, as shown in Fig.4a, the simulation successfully replicates the composition of crashes across various road geometries and crash types, closely mirroring real-world distributions from the city of Ann Arbor, Michigan[38]. Moreover, our simulations capture a broad range of urban traffic conditions, including variations in traffic density, speed, and signalized travel delay, as illustrated in Fig. 4b. Traffic density represents the number of vehicles present within the simulation environment, vehicle speed reflects the instantaneous speed of each vehicle, and traffic light waiting time captures the duration vehicles spent at signalized intersections. While we do not have real-world data for direct comparison, these results demonstrate the simulation's ability to generate diverse urban traffic patterns, highlighting its potential for comprehensive traffic representation.

Testing AVs directly within the NDE can be time-consuming due to the rarity of safety-critical events[22]. Studies suggest that hundreds of millions of miles of testing might be required to demonstrate AV safety performance at a human driver level[29]. The NADE amplifies the likelihood of safety-critical events by training BVs to execute strategic adversarial maneuvers that challenge the AV under test. This strategic approach significantly accelerates testing efficiency by orders of magnitude while ensuring that safety assessments remain unbiased[27]. By deploying the AV in the NADE and conducting a large number of runs, the DIT provides a comprehensive statistical assessment of AV safety, including crash rates, crash types, and crash severity statistics. Beyond efficiency, DIT offers a unique advantage in uncovering unknown unsafe behaviors. Its interactive, large-scale, and dynamic testing environment continuously generates safety-critical situations tailored to the specific AV under test. These emergent interactions naturally reveal failure modes that may remain hidden in static or scenario-based evaluations. As a result, DIT not only provides rigorous statistical validation but also serves as a powerful tool for exposing latent vulnerabilities—delivering a comprehensive assessment of AV driving intelligence.

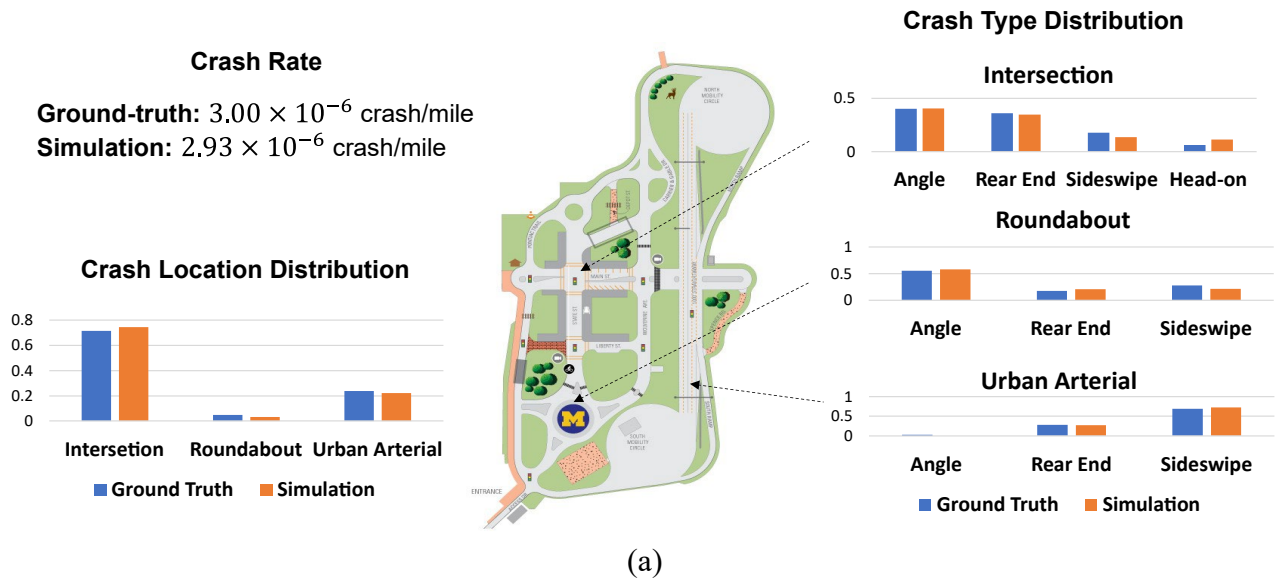

(a)

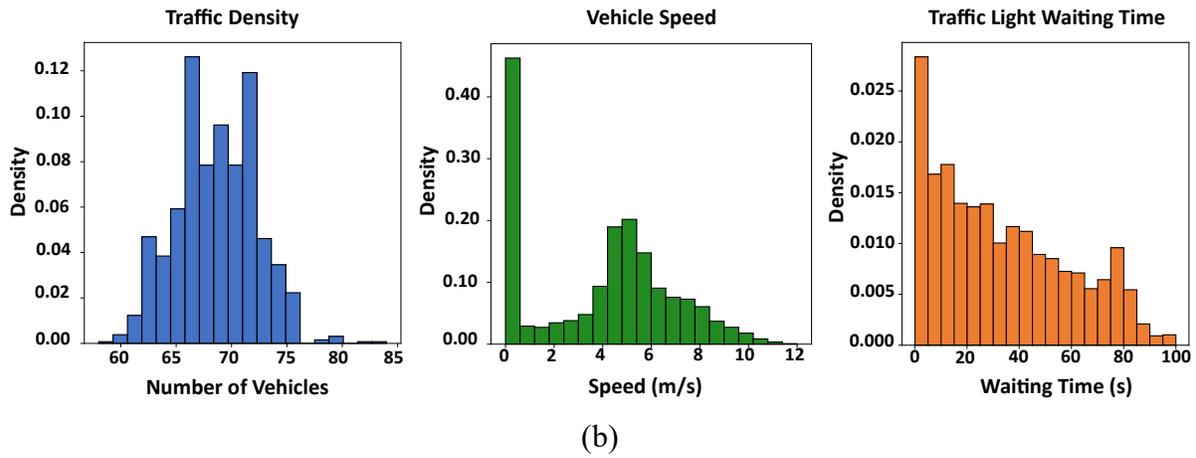

(b)

**Figure 4. Demonstration of the simulated Mcity environment for the Driving Intelligence Test. a,** Validation of the safety-critical events statistics. **b,** Illustration of the normal driving events statistics.

## AV under test

In this study, we aim to validate the proposed methodology by assessing the behavioral safety of AVs in urban driving environments. The AV's ODD is typical urban conditions in U.S. cities, with a speed limit of 30 mph, during daytime hours, and in sunny or cloudy weather. For our evaluation, we utilize an open-source Level 4 AV system: Autoware.Universe[39], the latest version of the widely adopted Autoware platform. Although its performance may not yet match that of commercial systems like Waymo, open-source platforms such as Autoware.Universe are among the most accessible and widely used Automated Driving System (ADS) available to the research community.

The evaluation was conducted at the Mcity[1] testing facility at the University of Michigan, Ann Arbor. We began by assessing the performance of Autoware.Universe within the simulated Mcity environment. To further demonstrate the real-world applicability of our framework, we extended the evaluation to two real vehicles: a Tesla Model 3 equipped with Full Self-Driving (version: v12.5.5) and a Lincoln MKZ equipped with Autoware.Universe. Compared to simulations, real-vehicle testing provides higher fidelity by incorporating factors such as vehicle dynamics and other physical effects that are often challenging to replicate in simulation.

## AV testing results in simulation

**Results of the Driver Licensing Test.** The overall test results of Autoware.Universe in the DLT are summarized in Table.1. Autoware.Universe successfully passed 6 out of 14 scenarios. The numbers of passed and failed cases, denoted as $n_p$ and $n_f$, are presented in the third and fourth columns of Table 1, respectively. Detailed performance metrics for each scenario—such as the average minimum time-to-collision ($\overline{TTC_{min}}$)—are analyzed only for scenarios in which Autoware.Universe achieved a passing result. Examples of both passed and failed scenarios are illustrated in Supplementary Movie 2.

**Table 1. The Driver Licensing Test results of Autoware.Universe.** The "P/F" column indicates whether Autoware.Universe passed (P) or failed (F) each scenario. A dash ("–") signifies that the corresponding metric was not computed due to failure in that scenario. "N/A" indicates that the metric is not applicable to the scenario.

| Scenario | N | P/F | $n_p$ | $n_f$ | $\overline{d_{min}}$ (m) | $\overline{TTC_{min}}$ (s) | $\overline{t_{react}}$ (s) |
|---|---|---|---|---|---|---|---|
| a. Cut-In | 46 | F | 32 | 14 | - | - | - |
| b. Car-Following | 40 | P | 40 | 0 | 5.51 | 3.61 | 3.55 |

---

[1] https://mcity.umich.edu/

| | | | | | | | |
|---|---|---|---|---|---|---|---|
| c. Lane Departure (same direction) | 46 | F | 39 | 7 | - | - | - |
| d. Lane Departure (opposite direction) | 51 | F | 0 | 51 | - | - | - |
| e. Left Turn (AV goes straight) | 57 | F | 36 | 21 | - | - | - |
| f. Left Turn (AV turns left) | 53 | P | 53 | 0 | 2.28 | 5.65 | 0.24 |
| g. Right Turn (AV goes straight) | 72 | F | 33 | 39 | - | - | - |
| h. Right Turn (AV turns right) | 37 | F | 26 | 11 | - | - | - |
| i. VRU Crossing the Street without the Crosswalk | 33 | F | 7 | 26 | - | - | - |
| j. VRU Crossing the Street at the Crosswalk | 50 | P | 50 | 0 | 3.46 | 2.02 | 1.31 |
| k. AV Merging into the Roundabout | 40 | P | 40 | 0 | 2.05 | 1.92 | 1.02 |
| l. BV Merging into the Roundabout | 74 | F | 19 | 55 | - | - | - |
| m. Vehicle Encroachment | 84 | P | 84 | 0 | 4.05 | 2.1 | N/A |
| n. Traffic Signal | 40 | P | 40 | 0 | N/A | N/A | N/A |

The testing results revealed several deficiencies in Autoware.Universe's basic behavioral competence, including failure to respond appropriately in the following scenarios: (1) a vehicle cutting in; (2) a vehicle departing its lane; (3) an oncoming vehicle drifting into the lane; (4) a left-turning vehicle while the AV proceeds straight at an intersection; (5) a right-turning vehicle while the AV proceeds straight; (6) a straight-moving vehicle while the AV turns right; (7) a VRU crossing the street without a crosswalk; and (8) a vehicle merging into a roundabout. The causes of collisions vary across scenarios. Two representative failure cases are shown in Fig. 5. In Case 30 of the *Left Turn (AV goes straight)* scenario (Fig. 5a), the AV begins close to the BV (Fig. 5a.2), requiring a deceleration greater than $-4.3$ m/s$^2$ to avoid a collision. Although Autoware.Universe initiates braking within 1 second (Fig. 5a.3), it fails to decelerate adequately. The peak deceleration reaches only $-3$ m/s² (Fig. 5a.1), resulting in a collision (Fig. 5a.4). In Case 8 of the *VRU Crossing Without Crosswalk* scenario (Fig. 5b), the initial distance between the AV and the pedestrian is 14.7 meters (Fig. 5b.2), indicating a manageable, low-risk situation. However, the Autoware.Universe system delays braking until the AV is very close to the VRU (Fig. 5b.3). Combined with a peak deceleration of only $-2.5$ m/s² (Fig. 5b.1), the delayed and weak response leads to a collision with the VRU (Fig. 5b.4).

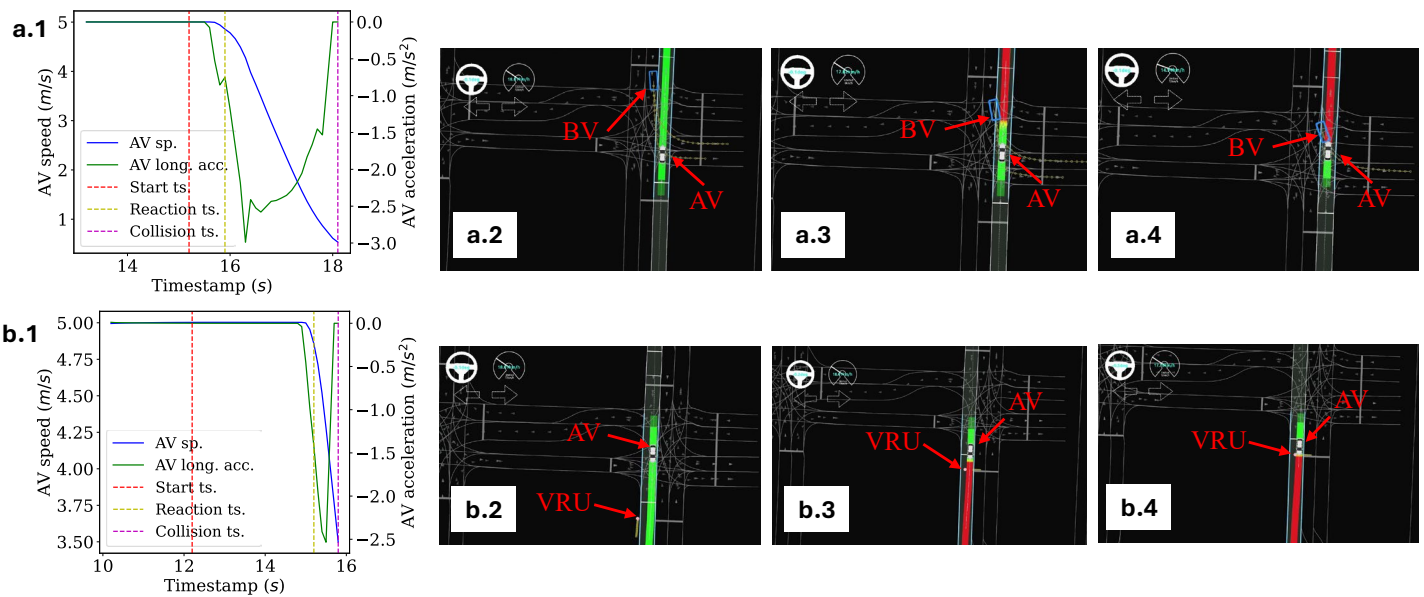

**Figure 5. Demonstration of failed cases in the Driver Licensing Test.** The white vehicle represents the AV, the blue-boxed vehicle represents the BV, and the pink agent represents the VRU. **a,** Case 27 in left-turn (AV goes straight) scenario, including the AV speed and acceleration curves during the test (a.1), where "sp." stands

for speed, "long." stands for longitudinal, "acc." stands for acceleration, and "ts." stands for timestamp, scenario beginning moment (a.2), the moment when Autoware.Universe starts to react (a.3), and the collision moment (a.4). **b,** Case 8 in VRU crossing the street without a crosswalk scenario, including the AV speed and acceleration curves during the test (b.1), scenario beginning moment (b.2), the moment when Autoware.Universe starts to react (b.3), and the collision moment (b.4).

For the scenarios that Autoware.Universe successfully passed, we further analyzed its performance using several metrics: average minimum distance ($\overline{d_{min}}$[2]), average minimum TTC ($\overline{TTC_{min}}$[3]), and average reaction time ($\overline{t_{react}}$[4]), as reported in Table 1. Some values are missing due to scenario characteristics where these metrics are not applicable. Across the passed scenarios, the average minimum distance ranges from 2.0 to 5.5 meters, and the average minimum TTC ranges from 1.9 to 5.7 seconds, suggesting that Autoware.Universe generally maintains safe following distances. In the *Left Turn (AV turns left)* scenario, the reaction time is notably low at 0.24 seconds, as the AV typically slows down in advance. In contrast, reaction times in the *AV Merging into the Roundabout* and *VRU Crossing at Crosswalk* scenarios are 1.02 and 1.31 seconds, respectively—longer than those observed in 75% of human drivers[40]. Autoware.Universe successfully avoids collisions with VRUs crossing at designated crosswalks but fails to recognize pedestrian intent when the crossing occurs outside of crosswalks. It also demonstrates the ability to avoid vehicles stopped at various angles on the road and to comply with traffic signals. These results demonstrate the capabilities of the DLT for evaluating AV basic behavior competencies.

**Results of the Driving Intelligence Test.** The Mcity testing environment and the AV's route for the DIT are depicted in Fig. 6a. The route incorporates various urban driving layouts, including urban arterials, signalized intersections (covering straight paths, unprotected left turns, and right turns), as well as single-lane and double-lane roundabouts. A full lap around the route constitutes a testing episode, which is reset if the AV either crashes or successfully completes the lap. The environment is stochastic, with randomized behavior and initialization of BVs in each lap, ensuring the AV encounters a wide range of real-world driving situations, as validated in previous sections. This dynamic setup mitigates the risk of gaming the test, ensuring a robust and reliable evaluation of the AV's performance. To evaluate AV's behavioral safety, we conducted around 3,500 episodes of testing around 936 wall-time hours with multiple AWS cloud server EC2 C5.9xlarge instances (each instance includes 36 cores and 72 GB RAM). Various types of crashes were observed throughout the testing route, examples of which are shown in Fig. 6b, including head-on (Fig. 6b.1), sideswipe (Fig. 6b.2), angle (Fig. 6b.3), and rear-end (Fig. 6b.4) collisions. The crash rate for Autoware.Universe was estimated at $3.01 \times 10^{-3}$ crashes per mile (Fig. 6c), about 1,000 times higher than that of an average human driver[38] ($3.00 \times 10^{-6}$ crash per mile). Detailed analyses of specific Autoware.Universe issues identified during the tests are discussed in the following paragraphs. Figs. 6d-e show the distribution of crash types and crash severity, which is classified into four levels based on the speed difference at the moment of impact, following existing literature[23]: level 1 (0-5 mph), level 2 (5-10 mph), level 3 (10-15 mph), and level 4 (above 15 mph). These results highlight the DIT's capacity to statistically evaluate the frequency of safety-critical events, providing a comprehensive assessment of the AV's safety performance under real-world conditions. A demonstration of DIT procedure can be found in Supplementary movie 3.

---

[2] The average value of the minimum distance between the AV and BV in each test case.
[3] The average value of the minimum TTC in each test case.
[4] The average reaction time of the AV in each test case. The reaction time means the duration between the case start moment and the moment when the AV start to react (e.g., decelerate, accelerate, and steering).

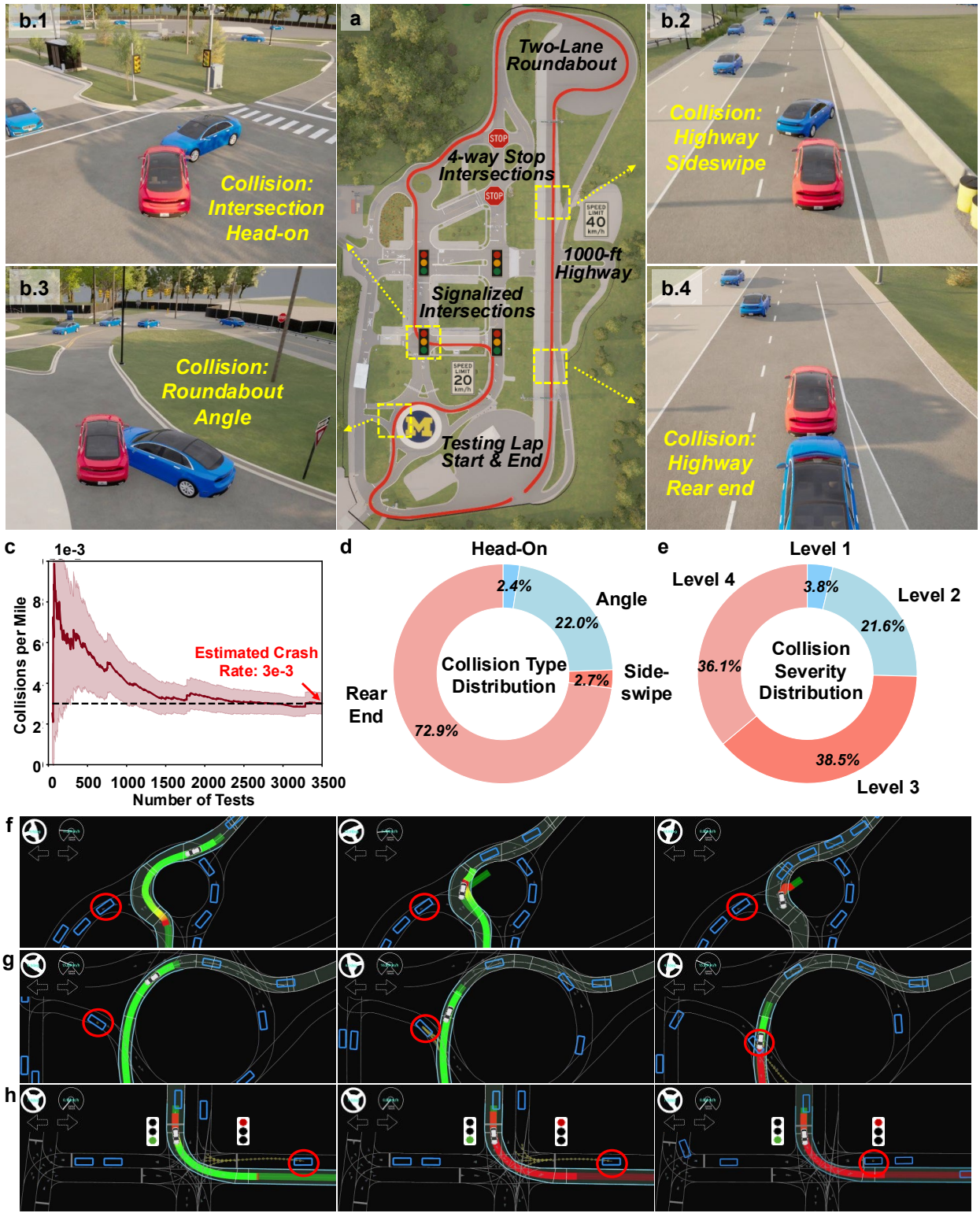

Figure 6. The Driving Intelligence Test results of Autoware.Universe. **a,** Demonstration of the Mcity testing environment and testing route of AV. **b,** Examples of different crash types experienced by Autoware.Universe: head-on (b.1), sideswipe (b.2), angle (b.3), and rear-end (b.4) collisions. The red vehicle represents the AV under test, while the blue vehicles represent background traffic. **c,** Crash rate estimation. **d,** Crash type estimation. **e,** Crash severity estimation. **f-h,** Demonstration of identified issues in Autoware.Universe: obstacle avoidance replanning error (f), right-of-way yielding issue (g), and trajectory prediction error (h). The white vehicle represents the AV under test, blue vehicles indicate background traffic, and the vehicle circled in red highlights the BV that is in conflict with the AV. A video demonstration can be found in Supplementary movie 4.

A key advantage of environment-level testing in the DIT is its continuity in time and space, with a large number of background agents, exposing the AV to a virtually infinite variety of potential driving situations. In this dynamic environment, the DIT can uncover hidden issues that are difficult to detect through scenario-based testing. We present three examples to demonstrate DIT's capability in identifying potential issues in Autoware.Universe. Notably, the AV is treated as a black box, meaning no prior knowledge of the AV is used in designing the testing environment or identifying problems. A video demonstration can be found in Supplementary movie 4. The first example, illustrated in Fig. 6f, involves the AV navigating a small roundabout in the northern section of Mcity (Fig. 6f left). When a BV slightly encroaches while entering the roundabout (as indicated by the red-circled vehicle), the AV attempts to replan its path, adjusting its trajectory to avoid close proximity to the BV (Fig. 6f middle). However, this replanning triggers an unknown error, causing the AV to become stuck at the replanned position (Fig. 6f right). This issue only occurs when the AV and BV occupy a specific, subtle relative position—not every time a BV merges—making it challenging to detect through modular testing or scenario-based approaches. This highlights the advantage of DIT's continuous, dynamic environment in uncovering unknown unsafe events. We reported this issue to the Autoware development team, who subsequently proposed a fix that successfully resolved the problem[5]. This example highlights the strength of our methodology in identifying hidden bugs in AV systems and supporting their development.

The second example highlights a right-of-way yielding issue observed in Autoware.Universe. When the AV has the right of way, it fails to proactively yield to aggressive BVs approaching from lower-priority lanes. Instead, the AV only responds once the BV enters its path. This behavior was especially problematic in scenarios like roundabouts (see Fig. 6g), where the AV was navigating while a BV attempted to merge. In Fig. 6g, the AV is shown driving in the roundabout (Fig. 6g left), the BV attempting to enter (Fig. 6g middle), and the resulting collision (Fig. 6g right). While it's understandable that the AV might choose not to yield in these cases as it has the right-of-way, this rigid behavior led to multiple crashes during testing. When faced with aggressive BVs showing no intention to stop, the AV should exercise greater caution to reduce the risk of collision. This example underscores the DIT's strength in identifying areas for improvement, providing valuable insights for enhancing AV models. The third example illustrates a trajectory prediction module issue, shown in Fig. 6h. The AV is driving southbound at an intersection with a green light for a left turn (Fig. 6h left). However, it begins to decelerate (Fig. 6h middle) and comes to a stop, obstructing southbound traffic at the intersection (Fig. 6h right). Upon further analysis, this behavior was due to the AV yielding to a BV traveling eastbound (circled in red) from a perpendicular lane, predicting that the BV would continue at its current speed to pass through the intersection, despite the BV facing a red light. This issue stems from the trajectory prediction module not incorporating traffic light information, leading to an incorrect conflict prediction. This issue has been acknowledged as a known limitation in the Autoware.Universe manual[6], reinforcing the DIT's ability to uncover critical design flaws in AV behavior.

**Demonstration of AV testing in physical test tracks**

After thoroughly evaluating Autoware.Universe in simulation, we extended the experiment to assess real AV systems—specifically a Tesla Model 3 and a Lincoln MKZ—on Mcity physical test tracks. Given the resource constraints, the objective of these physical tests was not to generate comprehensive testing results but to validate the applicability of our framework in real-world closed testing facilities. By testing real vehicles, we aimed to demonstrate how the framework can be applied in actual testing procedures, facilitating third-party evaluation of AV systems. Demonstrations of testing events can be found in Supplementary Movie 5.

---

[5]GitHub link of the reported issue: https://github.com/autowarefoundation/autoware.universe/issues/5687
[6]https://autowarefoundation.github.io/autoware.universe/main/planning/behavior_velocity_planner/autoware_behavior_velocity_intersection_module/#how-towhy-set-rightofway-tag

## Demonstration of the Driver Licensing Test

The Driver Licensing Test was conducted at the Mcity test facility to evaluate Tesla's Full Self-Driving (FSD) feature, as shown in Fig. 7a.1. Two scenarios were tested: the left-turn (AV goes straight) and the VRU crossing the street without the crosswalk. To ensure safety during the testing process, Humanetics robotic platforms[7], specifically UFO Pro for BV (Fig. 7a.2) and UFO Nano for VRU (Fig. 7a.3), equipped with a dummy vehicle and child were used instead of a real vehicle and pedestrian. These platforms are specifically designed for vehicle testing, featuring a chassis that is sufficiently low to allow real vehicles to drive over them during collisions, while being robust enough to withstand the impact without sustaining damage. Each platform is equipped with a real-time kinematic (RTK) global positioning system (GPS). Similarly, an RTK system was installed on the AV to collect real-time state information during the test. The platform is controlled adaptively to satisfy the initial conditions of the testing case according to the AV and BV/VRU state information. Importantly, the RTK GPS is the only device attached to the AV under test to collect its state information for triggering the test scenario, and the testing system did not transmit any information to the AV. The AV relied solely on its own sensors and algorithms to perform driving tasks, serving as a black-box to the testers.

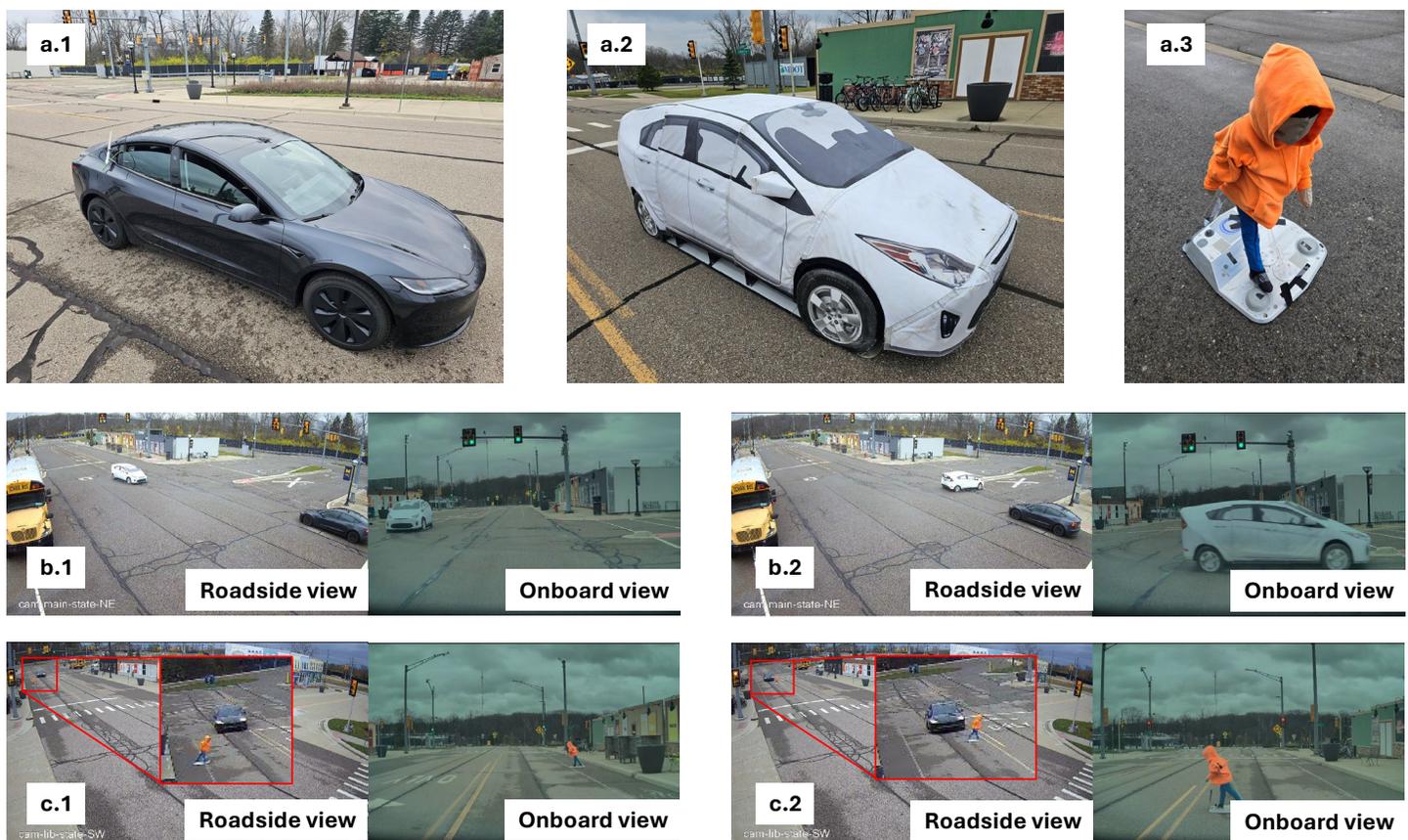

**Figure 7. The field experiment of the Driver Licensing Test. a.1-a.3,** The test system. **a.1,** The vehicle under test equipped with the RTK system. **a.2,** The Humanetics UFO Pro platform installed with the dummy vehicle. **a.3,** The Humanetics UFO Nano platform installed with the dummy child. **b.1-b.2,** Testing process for the left turn (AV goes straight) scenario. **b.1,** Images captured by the roadside camera and Tesla's front-facing camera at the scenario start moment, defined as the point when the relative distance and speed satisfy the conditions specified in the test case. **b.2,** Images captured by the roadside camera and Tesla's front-facing camera when the AV comes to a stop. **c.1-c.2,** Testing process for the VRU crossing the street without the crosswalk scenario. **c.1,**

---

[7] https://www.humaneticsgroup.com/products/active-safety-test-systems

Images captured by the roadside camera and Tesla's front-facing camera at the scenario start moment. **c.2,** Images captured by the roadside camera and Tesla's front-facing camera when the AV comes to a stop.

The measured free-driving speed of Tesla's FSD system was 8.94 m/s (20 mph) on the test road. Based on this speed, risk levels and test cases were generated. Considering time constraints and platform capabilities, five test cases were conducted for each scenario with at least one case for each risk level. Tesla's FSD system successfully passed all ten sampled test cases. For each scenario, we choose one test case and illustrate it in Fig. 7b and Fig. 7c, respectively. Figs. 7b.1-7b.2 demonstrate time moments when the case started and AV successfully stopped in the left-turn scenario, taken by the roadside and AV front-facing cameras. It can be found that Tesla can successfully identify potential dangerousness and apply reasonable behavior to stop and let the aggressive BV pass first. Similarly, Figs. 7c.1-7c.2 shows the Tesla can properly react to the VRU jaywalking situation, demonstrating behavior competency in such safety-critical scenarios that commonly occur in urban ODD. These experiments demonstrate that the proposed DLT enables third-party testing of commercial vehicles in physical test tracks. During the test, the AV is treated as a black box, with no reliance on any data application programming interfaces (APIs), demonstrating its effectiveness for third-party AV safety evaluation.

**Demonstration of the Driving Intelligence Test**

The Driving Intelligence Test was conducted at the Mcity test facility to assess the performance of Autoware.Universe using a real vehicle. The vehicle under test was a retrofitted Lincoln MKZ, equipped with the Dataspeed drive-by-wire system and a suite of sensors, including LiDAR, radar, camera, RTK GPS, and IMU. The vehicle's planned trajectory, generated by Autoware.Universe, was executed using Preview Control method[41], which provides control commands for throttle, brake, and steering angle. To streamline the testing process, we bypassed the perception module and fed ground-truth object states directly into the AV stack.

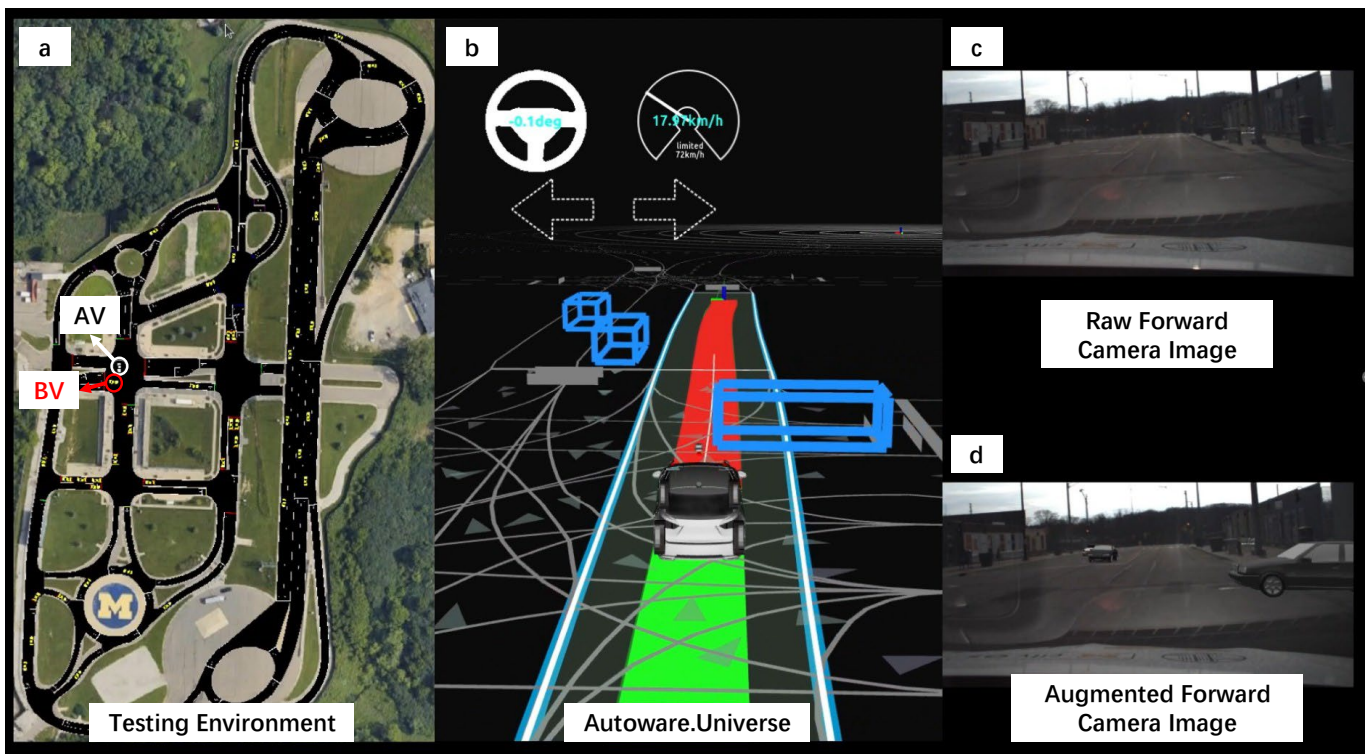

**Figure 8. A snapshot of the field experiment of the Driving Intelligence Test. a,** Testing environment view showing the complete traffic environment of Mcity, with the white vehicle representing the AV and all yellow vehicles representing the BVs. The red circled vehicle denotes the conflicting BV. **b,** Autoware.Universe's view,

with the white vehicle representing the AV and surrounding BVs indicated by blue boxes. **c,** Raw forward-facing camera image from the AV. **d,** Augmented forward-facing camera image, with BVs integrated into the scene.

Unlike the DLT, the DIT involves a larger number of background agents, making it difficult to use physical proxy robots as in the DLT. To address this challenge, we leveraged the Augmented Reality (AR) platform at the Mcity facility, where a simulated traffic environment is synchronized with the physical world, including the AV and traffic infrastructure (e.g., traffic signals). This setup enables bidirectional information flow between the virtual BVs and the real AV, allowing interaction between the two with only the AV present on the physical test track. This approach facilitates testing in a realistic traffic environment without requiring large numbers of real vehicles, reducing both costs and safety risks. Further details about the AR platform are provided in the Methods section.

The testing route and conditions were consistent with the simulation tests described earlier. We conducted multiple test laps and observed a variety of crash events, consistent with the simulation testing results. Due to space constraints, we present one example in Fig. 8. This figure shows a snapshot from the field experiment, where Fig. 8a depicts the complete traffic environment at Mcity, with the white vehicle representing the AV and the yellow vehicles representing BVs. Fig. 8b shows the AV's view, with surrounding BVs highlighted by blue boxes. Figs. 8c and 8d display raw and augmented images from the AV's forward-facing camera, respectively. In this example, the AV is traveling southbound through a signalized intersection with a green light, while a BV runs a red light from the west. The AV decelerates but ultimately collides with the BV. Although the AV had the right-of-way, a more proactive safety maneuver could have been executed to avoid the collision, resulting in better overall safety performance. Using the AR platform, we can either feed the augmented images directly into the AV's sensing system to evaluate the full AV pipeline, or send object information to focus specifically on testing the decision-making component of the AV. A video version of this scenario, along with additional testing scenarios, can be found in Supplementary Movie 5. These results demonstrate the capability of the proposed method for evaluating AVs in real-world physical test tracks.

## Discussion

In this study, we present a systematic and standardized third-party testing framework for assessing AV behavioral safety, operating under a black-box approach that requires no prior knowledge of the AV's internal workings. The framework aims to create a comprehensive blueprint that unifies testing objectives, procedures, and measurements with executable methodologies that can be refined and expanded to support large-scale AV deployment. This evaluation framework addresses both minimum performance requirements, such as behavioral competency, and comprehensive performance assessment, such as the occurrence rate of measurable safety-related events, like crash rates, to comprehensively evaluate AV behavior performance. The testing scenarios and environments are designed to be both naturalistic and adversarial, substantially improving testing efficiency without sacrificing unbiasedness. The proposed method is versatile, applicable to both simulation and physical test tracks, and has the potential to empower testing facilities worldwide in advancing AV behavioral safety testing and validation.

One of the most significant challenges in evaluating AV behavioral safety is bridging the sim-to-real gap, particularly for the DIT. This gap stems from two primary sources: the fidelity of simulated sensor inputs for the AV stack and the realism of human agents' behavior in simulations. In our current case studies with Autoware.Universe, we simplified the testing process by bypassing the perception module and feeding object ground-truth states directly into the AV stack. As end-to-end AV stacks become increasingly prevalent, our testing framework can be extended to incorporate state-of-the-art sensor simulation packages, such as NVIDIA's Omniverse and Cosmos. Additionally, achieving high-fidelity simulations that accurately replicate human driving behavior over extended time period remains a significant challenge. Recent advancements, such as the development of world models, have shown potential in addressing these issues, but continued research and innovation are necessary to fully close these gaps and ensure the reliability of AV testing.

Behavioral compliance with traffic regulations is fundamental to AV safety and should be evaluated. In the DLT, diverse scenarios should be designed to assess whether AVs adhere to traffic laws. In this study, we only demonstrate it with a single case with traffic light, evaluating whether AV can properly stop at red lights. More scenarios should be designed and evaluated according to the local traffic laws and regulations. Additionally, in the DIT, compliance can be monitored dynamically, for example, using trigger-based violation detection methods proposed in literature[42] to identify infractions in naturalistic driving environments. Beyond safety, AV behavior also impacts broader traffic dynamics, including mobility and congestion. For instance, Cruise AVs in San Francisco were reported to have blocked intersections, causing severe traffic disruptions[43]. Additionally, AVs may adopt different car-following headways compared to human drivers, potentially impacting overall traffic flow and efficiency. While this paper focuses on safety and mobility and congestion effects are out of the scope of this study, dedicated evaluation frameworks for these secondary effects should be developed in the future, as they also play a crucial role in the large-scale deployment and public acceptance of AVs.

## Methods

**Formulation of AV behavioral safety assessment problem**

This section describes the problem formulation of AV behavioral safety assessment problem. Denote the variable of the driving environment as $\mathbf{x} = [\mathbf{s}(0), \mathbf{u}(0), \mathbf{u}(1), ..., \mathbf{u}(T)]$, in which $\mathbf{s}(k)$ denote the state vector (position, speed, heading, etc.) of the AV and BVs at the $k$-th timestep, and $\mathbf{u}(k)$ denote the action vector of BVs at this step. The core objective of behavioral safety evaluation is to determine whether an AV system can operate safely within this dynamic driving environment. Let $\pi$ denote the AV model and its safety performance $\sigma_\pi$ can be measured by

$$\sigma_\pi = \mathcal{F}(\mathbf{x}, \pi), \mathbf{x} \sim P(\mathbf{x}),$$

where $\mathcal{F}(\cdot)$ denotes testing methods and procedures that derive AV safety performance from the driving environment $\mathbf{x}$, and $P(\mathbf{x})$ denotes the distribution of the NDE.

Behavioral safety is dependent on two critical aspects: (1) the AV's behavioral capability when confronted with potentially hazardous situations, and (2) the AV's ability to avoid entering such dangerous situations altogether. The first aspect addresses the AV's performance in safety-critical events triggered by itself or other agents, such as a sudden cut-in by an adjacent vehicle. The second aspect evaluates the AV's overall competence of navigating the environment. It involves not only the capability of reacting to dangerous situations but more importantly its capabilities of reducing the chances of posing itself in dangerous situations, for example, not staying in blind spot of other vehicles to reduce the chances of being cut in.

To comprehensively evaluate these two aspects, we propose a dual testing framework consisting of the DLT and the DIT. In the DLT, the initial condition of the driving environment $\mathbf{s}(0)$ and the behavior of BVs $\mathbf{u}(k)$ are predefined to establish a specific testing scenario. The testing result for each testing episode is binary—whether a safety-critical event occurs or not

$$\sigma_\pi(\mathbf{x}) = \mathbb{I}(A|\mathbf{x}, \pi), \mathbf{x} \sim P(\mathbf{x}),$$

where $A$ denote the safety-critical event, for example, crash event is used in this study, $\mathbb{I}(\cdot)$ is the indicator function which equals to one if a crash happens during the testing case $\mathbf{x}$ and equals zero otherwise, and $P(\mathbf{x})$ here denotes the distribution of testing cases in the scenario set. Based on the pass and fail results of AV under diverse testing cases, we can evaluate its behavior competence of handling dynamic driving tasks within the ODD.

The Driving Intelligence Test is designed to assess statistical safety performance of AV, which can be measured by

$$\sigma_\pi = P(A) = \mathbb{E}(P(A|\mathbf{x})), \mathbf{x} \sim P(\mathbf{x}),$$

where $P(A)$ denotes the crash rate of AV. The testing process is to deploy the AV in the simulated NDE, observe its safety performance during the dynamic interaction with background agents, and obtain the occurrence rate of safety-critical events by accumulating a large amount of testing episodes. This evaluated crash rate demonstrates the overall safety performance of AVs and provides statistical evidence of AV behavioral safety during real-world traffic deployments.

As safety-critical events are rare events in the real-world traffic environment, most driving scenarios and environments **x** in both DLT and DIT would be not challenging for AV. As a result, it will take large number of tests to obtain the evaluation results[23,29], which is time-consuming and even impractical for real-world tests. To improve testing efficiency, the key is to generate a new testing environment distribution $q(\mathbf{x})$, we termed it Naturalistic and Adversarial Driving Environment (NADE), in which safety-critical events are intelligently identified and generated to increase the exposure of AV to these high-valued events. Therefore, for both DLT and DIT, the testing environment will be sampled from the NADE rather than NDE, aiming to evaluate the AV behavioral safety with high accuracy and efficiency.

To summarize, the DLT and the DIT follow the same overall framework, first constructed the naturalistic environment distributions $P(\mathbf{x})$ from real-world datasets, then generated an intelligent environment distribution $q(\mathbf{x})$ to improve testing efficiency. The DLT is a simplified case of the DIT, with a smaller number of background agents and shorter time intervals. The detailed processes of constructing both tests are discussed in the following sections.

**Driver licensing test**

The DLT is developed in four steps: scenario selection, scenario modeling, scenario calibration, and scenario case generation. The selection of test scenarios depends on the ODD of the AV deployment area. In this study, we demonstrate the evaluation of the AV in urban ODD. Based on the testing library proposed in Ref[34], we identified 14 scenarios, which constitute the testing scenario set, as illustrated in Fig.3a. The black vehicle in the figure demonstrated the AV, and the white vehicle and human icon demonstrated the background agent in the scenario. Note that we are not arguing that the 14 scenarios are comprehensive in terms of coverage for urban ODD, and we mainly use them to demonstrate the proposed methodology with the consideration of time and labor budget. The proposed method is appliable for different scenario libraries and more testing scenarios.

The second step is scenario modeling, in which we parameterize each scenario and construct it in the state space **x**. The dynamic variables for each scenario are the initial relative position and speed between AV and BVs. For example, the cut-in scenario is modeled as a two-dimensional problem including the relative speed and distance between AV and cut-in BV. After initializing the desired traffic situation, the behavior of BVs is pre-determined and will not react to the behavior of AV for easy execution in physical tests. In this study, BVs will maintain a constant speed during the scenario. The parameterization of dynamic variables for each scenario can be found in Supplementary Materials.

The third step is scenario calibration, determining the distribution of dynamic variables based on human driving data to construct naturalistic distribution $P(\mathbf{x})$. This step is important to ensure that the scenarios are representative of situations that AV will face during deployment, and to rule out infeasible parameter settings that the collision is unavoidable with BVs, or the behavior exceeds AV physical limits. We leverage naturalistic driving data (NDD) to extract vehicle trajectories for calibrating each scenario. More details on the scenario modeling and calibration can be found in Supplementary Materials.

The fourth step is scenario case generation, determining concrete test cases of each scenario. Instead of randomly sampling from $P(\mathbf{x})$ based solely on occurrence rates, we want to balance the coverage and the difficulty of test cases to cover different risk levels. To achieve this, we first determine if a case is hard or not by calculating the required minimum deceleration/acceleration of the AV to avoid the collision. The larger the value is, the riskier the case. Based on the value and NDD, we divide cases into five different risk levels, that is, trivial, low, medium, high, and infeasible risk level. To ensure the coverage of the test case space, we leverage the covering suites to sample the test[44]. For each risk level—excluding the trivial level, where test cases are considered very safe and require minimal or no action from the vehicle, and the infeasible level, where collisions are unavoidable—we discretized the test case state space into cubes, and each cube center parameter serves as a test case. As a result, the state space segmented with risk levels constitute the new sampling distribution $q(\mathbf{x})$, as shown in Fig.3b (right), in which covering the state space and diverse risk levels.

**Driving intelligence test**

The DIT aims to evaluate the crash rate of AV, which can be obtained by the Monte Carlo method[45]

$$P(A) = \mathbb{E}_{\mathbf{x} \sim P(\mathbf{x})}\big(P(A|\mathbf{x})\big) \approx \frac{1}{n} \sum_{i=1}^{n} P(A|\mathbf{x}_i), \mathbf{x}_i \sim P(\mathbf{x}),$$

where $n$ is the number of testing episodes, and $i$ denotes the $i$-th test. The actions of BVs $\mathbf{u}(k)$, conditioned on the current state $\mathbf{s}(k)$, follow a realistic distribution $P\big(\mathbf{u}(k)|\mathbf{s}(k)\big)$. With the Markovian assumptions of BV behavior, the naturalistic probability of each testing episode can be calculated as $P(\mathbf{x}) = P\big(\mathbf{s}(0)\big) \times \prod_{k=0}^{T} P\big(\mathbf{u}(k)|\mathbf{s}(k)\big)$. To simulate the NDE in Mcity, we adopted TeraSim[36], a SUMO[46]-based, open-source, high-fidelity traffic simulation platform designed specifically for AV safety testing. The proposed NDE environment has been validated to produce realistic crash rates, crash type distributions, and plausible traffic conditions, as illustrated in Figs. 4a-b.

To improve testing efficiency, we developed the NADE based on NDE, forming an intelligent testing environment $q(\mathbf{x})$ to evaluate AV performance based on importance sampling theory[45]

$$P(A) = \mathbb{E}_{\mathbf{x} \sim q(\mathbf{x})} \big(P(A|\mathbf{x}) \cdot W_q(\mathbf{x})\big) \approx \frac{1}{n} \sum_{i=1}^{n} P(A|\mathbf{x}_i) \cdot W_q(\mathbf{x}_i), \mathbf{x}_i \sim q(\mathbf{x}),$$

where $W_q(\mathbf{x})$ denotes the weight of each testing episode, i.e., *likelihood ratio*, ensuring the unbiasedness of testing results. It can be calculated by

$$W_q(\mathbf{x}) = \frac{P(\mathbf{x})}{q(\mathbf{x})} = \prod_{k=0}^{T} \left[\frac{P\big(\mathbf{u}(k)|\mathbf{s}(k)\big)}{q\big(\mathbf{u}(k)|\mathbf{s}(k)\big)}\right],$$

where $q\big(\mathbf{u}(k)|\mathbf{s}(k)\big)$ denote BVs behavior distribution in the NADE.

To develop the NADE, we applied methods proposed in Ref[27]. The core idea is to identify the critical BV, referred to as the Principal Other Vehicle (POV), and adjust its behavior to challenge the AV. To identify the POV, we define the criticality value for the $j$-th BV at the $k$-th timestep as

$$C_j(\mathbf{s}(k)) = \sum_{\mathbf{u_j}(k)} V\left(\mathbf{u_j}(k)|\mathbf{s}(k)\right),$$

where $\mathbf{u_j}(k)$ denotes the $j$-th BV's action, and $V\left(\mathbf{u_j}(k)|\mathbf{s}(k)\right)$ denotes the maneuver criticality of the action, calculated as

$$V\left(\mathbf{u_j}(k)|\mathbf{s}(k)\right) = P\left(\mathbf{u_j}(k)|\mathbf{s}(k)\right) \cdot P\left(A_j|\mathbf{s}(k), \mathbf{u_j}(k)\right),$$

where $A_j$ denotes the crash between the $j$-th BV and AV. The maneuver criticality value is the product of the BV's action exposure frequency $P\left(\mathbf{u_j}(k)|\mathbf{s}(k)\right)$, which comes from the NDE model, and the corresponding maneuver challenge $P\left(A_j|\mathbf{s}(k), \mathbf{u_j}(k)\right)$, which estimates the likelihood of a crash based on calibrated human driving behavior as a surrogate for the AV.

At each timestep, if at least one BV has a criticality value greater than zero, we define it as a critical moment, and the BV with the highest criticality is identified as the POV. To accelerate testing, all other BVs will maintain naturalistic behavior while the POV's behavior is intelligently modified. The importance function $q(\mathbf{u}(k)|\mathbf{s}(k))$ is developed as

$$q(\mathbf{u}(k)|\mathbf{s}(k)) = q\left(\mathbf{u_j}(k)|\mathbf{s}(k)\right) \cdot \prod_{m=1, m \neq j}^{N} P(\mathbf{u_m}(k)|\mathbf{s}(k)),$$

where $M$ represents the number of BVs. Specifically, the POV's adversarial maneuver probability is amplified by constructing its new behavior distribution as a weighted average of the naturalistic and normalized criticality distributions

$$q\left(\mathbf{u_j}(k)|\mathbf{s}(k)\right) = \epsilon \cdot \frac{V\left(\mathbf{u_j}(k)|\mathbf{s}(k)\right)}{C_j(\mathbf{s}(k))}.$$

Here, $\epsilon$ balances the exploration and exploitation of adversarial maneuvers, mitigating the impact of criticality approximation errors introduced by the surrogate model. In this study, $\epsilon$ is set to 3000, in line with the configuration in Ref[27]. Theoretical analysis in Ref[27] has also demonstrated that testing AVs using the proposed NADE framework is both efficient and unbiased, offering a solid theoretical foundation for the methodology.

**Simulation settings**
**NDE simulator for Driving Intelligence Test.** To simulate NDE in Mcity, Intelligent Driver Model (IDM)[47] car-following and LC2013[48] lane-changing models are utilized in the TeraSim. For the IDM model, two sets of model parameters from Refs[49,50] that are calibrated using the Waymo Open Motion Dataset[51] and the 100-Car Naturalistic Driving Study[52] are used to simulate US urban driving environments. Detailed model parameters can be found in Supplementary Materials. For the LC2013 model, we employ the default parameter settings developed by SUMO[46]. However, simply using car-following and lane-changing models cannot generate realistic safety-critical events, as they are designed for crash-free situations. Therefore, to generate safety-critical events, we developed the human error model, a simple yet effective approach inspired by empirical studies[57] on the critical reasons of vehicular crashes, which identify the recognition error due to human inattention as the major factor in real-world collisions. This model operationalizes these insights by assigning each simulated vehicle a probability of neglecting surrounding objects—such as other vehicles, traffic lights, and traffic signs—for a certain duration at each timestep. This probabilistic approach enables the generation of a wide spectrum safety-critical scenarios, including but not limited to highway cut-ins, fail-to-yield at roundabout, and intersection conflicts such as unprotected left turns and red-light running incidents. More details for safety-critical events generation can be found in Supplementary Materials. Leveraging 5 years (2016-2021) crash data from Michigan, US[38], the human error source probabilities are calibrated, ensuring the consistency of crash rate, distribution of simulated crash types and locations with empirical crash statistics, as illustrated in Fig.4a. The proposed DIT methodology is compatible with various NDE approaches, and the model-based NDE simulation method can be further enhanced by incorporating recent advances in data-driven techniques, which could further improve simulation fidelity.

**NADE environment for Driving Intelligence Test.** The NADE environment for the DIT builds upon the methodology presented in our previous work[27], with several enhancements. To improve its efficiency, we employ a selective amplification method for collision probabilities based on location. According to real-world data, collisions are distributed unevenly across different locations; for example, the majority occur at intersections, while collisions at roundabouts account for less than 10% of the total. Assigning equal importance sampling probabilities to collisions at different locations would result in an imbalance and require longer episodes to achieve a stable estimation of the overall collision rate. To address this, we multiply the base probability of each predicted collision type by a fixed factor of $\epsilon = 3000$. To ensure a balanced generation of diverse accidents, we cap the amplified value using different upper bounds for different types. In this study, we use 0.1 for intersection collisions and 0.01 for other types, such as urban arterial or roundabout collisions. During the test, effective episodes exclude cases where the simulation or AV system behaves abnormally, for example, AV getting stuck on the test route.

**Physical test settings**

**Humanetics testing platform.** The Humanetics testing system consists of the robotic platform, dummy vehicle/VRU, vehicle under test system (i.e., AV), base station, control panel, and a computer. The dummy vehicle/VRU is installed on the robotic platform, which is equipped with an RTK GPS, serving as the background agent. There is also an RTK GPS that can be attached on the AV under test. Both the robotic platform and the AV can receive the RTK correction information from the base station to obtain highly accurate localization. The platform can also receive the AV state information through the base station to adjust its behavior to satisfy the initial condition of the test case. The control panel can control the platform to start and end the test. Dedicated software provided by Humanetics can be used to set up the test case and record the localization, speed, heading, etc. of both the platform and the AV through communication with the base station.

**Augmented reality testing platform.** To enable the testing of real AVs in physical test tracks, we developed the AR testing platform that seamlessly integrates and synchronizes a virtual, simulated traffic environment with the physical testing environment. This platform facilitates real-time bidirectional communication between the real-world AV and virtual BVs through Redis[8], an open-source in-memory data store used as a message broker. Information about virtual BVs, including identifiers, location, speed, and other relevant data, is transmitted to the AV on the test track. Simultaneously, the state of the AV is synchronized back to the virtual environment, ensuring realistic interactions between the AV and BVs. Additionally, virtual agents are rendered and blended into the AV's camera feed using pyrender[53], which serves as sensor input for the AV system. More details of the platform can be found in Refs[54,55].

**Data Availability**

The raw datasets that we used for modeling the naturalistic driving environment in Driver Licensing Test come from Argoverse 2 Motion Forecasting Dataset[35] and rounD Dataset[56]. The normal driving environment parameters in the Driving Intelligence Test are sourced from two calibrated models from Refs[49,50] based on the Waymo Open

---
[8]https://redis.io/

Motion Dataset[51] and the 100-Car Naturalistic Driving Study[52]. The safety-critical events generation parameters in Driving Intelligence Test are calibrated using data from Michigan Traffic Crash Facts[38]. All experimental data of this study are available at: https://zenodo.org/records/15446739.

**Code Availability**

The simulation software SUMO is publicly available, as described in the text and the relevant references[46]. All source codes of this study, including the Driver Licensing Test, the Driving Intelligence Test, and the customized automated driving system Autoware.Universe are available at: https://zenodo.org/records/15446750.

## Acknowledgments

This research was partially funded by the U.S. Department of Transportation (USDOT) Automated Driving System Demonstration Grant (#693JJ319NF00001), USDOT Region 5 University Transportation Center: Center for Connected and Automated Transportation (CCAT) of the University of Michigan (#69A3551747105), and the National Science Foundation (CMMI #2223517). Any opinions, findings, conclusions, or recommendations expressed in this material are those of the authors and do not necessarily reflect the official policy or position of the U.S. government.

We thank the late Prof. Huei Peng of the University of Michigan and Dr. Tony Geara of the City of Detroit for insightful discussions and thank Jim Lollar and Alex Rozelle of the University of Michigan for supporting field tests at the Mcity Test Facility.

## Author contributions

H. L. and X. Y. conceived and led the research project. H. L., X. Y., H. S., and T. W. developed the AV behavioral safety assessment framework and algorithm and wrote the paper. T. W. developed the Driver Licensing Test. H. S., H. Z., and X. Y. developed the Driving Intelligence Test. H. S., T. W., Z. Q., and H. Z. performed simulation experiments. X. Y., H. S., T. W., Z. Q., H. Z., and S. S. conducted field tests. All authors provided feedback during the manuscript revision and results discussions. H. L. approved the submission and accepted responsibility for the overall integrity of the paper.

## Competing interests

The authors declare no competing interests.

## Supplementary Movies

To better illustrate the methodology and results of our approach, we provide the following Supplementary Movies, which are available online at https://zenodo.org/records/15110539.
Supplementary movie 1: Illustration of the Driver Licensing Test.
Supplementary movie 2: Demonstration of passed and failed testing scenarios in the Driver Licensing Test.
Supplementary movie 3: Illustration of the Driving Intelligence Test.
Supplementary movie 4: Demonstration of three safety-critical cases identified in the Driving Intelligence Test.
Supplementary movie 5: Demonstration of AV testing at the Mcity physical test track.